\documentclass[lettersize,journal]{IEEEtran}
\usepackage{amsmath,amsfonts}
\usepackage{algorithmic}
\usepackage{algorithm}
\usepackage{array}
\usepackage[caption=false,font=normalsize,labelfont=sf,textfont=sf]{subfig}
\usepackage{textcomp}
\usepackage{stfloats}
\usepackage{url}
\usepackage{verbatim}
\usepackage{graphicx}
\usepackage{cite}
\usepackage{graphicx}
\usepackage{amsfonts}
\usepackage{booktabs} 
\usepackage{amsmath}
\usepackage{longtable}
\usepackage{amssymb}
\usepackage{multirow}
\usepackage[numbers,sort&compress]{natbib}
\usepackage{setspace}
\usepackage{color}

\begin{document}

\title{Rega-Net: Retina Gabor Attention for Deep Convolutional Neural Networks}

\author{Chun Bao, Jie Cao, Yaqian Ning, Yang Cheng, Qun Hao
	\thanks{This work was supported in part by the funding of foundation enhancement program under Grant 2019-JCJQ-JJ-273, in part by the National Natural Science Foundation of China under Grant 61871031 and Grant 61875012.{\em(Corresponding author: Jie Cao.)}}
	\thanks{The authors are with the Key Laboratory of Biomimetic Robots and Systems, Ministry of Education, Beijing Institute of Technology, Beijing 100081, China, and Chun Bao, Jie Cao, Yang Cheng, Qun Hao are also with Yangtze Delta Region Academy, Beijing Institute of Technology, Jiaxing 314003, China (e-mail: baochun@bit.edu.cn; caojie@bit.edu.cn; daydayupnyq@163.com; yangcheng2007@163.com; qhao@bit.edu.cn)}
}

\markboth{IEEE}
{Shell \MakeLowercase{\textit{et al.}}: Bare Demo of IEEEtran.cls for IEEE Journals}
\maketitle


\maketitle

\begin{abstract}
	Extensive research works demonstrate that the attention mechanism in convolutional neural networks (CNNs) effectively improves accuracy. Nevertheless, few works design attention mechanisms using large receptive fields. In this work, we propose a novel attention method named Rega-net to increase CNN accuracy by enlarging the receptive field. Inspired by the mechanism of the human retina, we design convolutional kernels to resemble the non-uniformly distributed structure of the human retina. Then, we sample variable-resolution values in the Gabor function distribution and fill these values in retina-like kernels. This distribution allows essential features to be more visible in the center position of the receptive field. We further design an attention module including these retina-like kernels. Experiments demonstrate that our Rega-Net achieves 79.96\% top-1 accuracy on ImageNet-1K classification and 43.1\% mAP on COCO2017 object detection. The mAP of the Rega-Net increased by up to 3.5\% compared to baseline networks.
\end{abstract}

\begin{IEEEkeywords}
	attention mechanism, retina-like kernels, Gabor
\end{IEEEkeywords}
\IEEEpeerreviewmaketitle
\vspace{-0.5cm}

\section{Introduction}

\IEEEPARstart{D}{eep}  learning networks are now being used in various fields related to computer vision, such as image recognition \cite{ref1,ref2}, object detection and recognition \cite{ref3,ref4,ref5}, image dehazing \cite{ref6,ref7}, and 3D vision \cite{ref8,ref9,ref10,ref11}. For image recognition deep learning networks, accuracy is one of the most critical evaluation metrics. There are various methods to improve the accuracy of neural networks, such as images or video pre-processing \cite{ref12,ref13}, hyper-parameters optimisation \cite{ref14,ref15}, or attention mechanisms. Attention mechanisms have been introduced into computer vision systems inspired by the human visual system. We generally put attention modules into the part of feature extraction to increase the network’s accuracy. In recent years, a great deal of work has been devoted to the design of attention modules for computer vision \cite{ref16,ref17,ref18}. `

\begin{figure}
	\centerline{\includegraphics[width=\columnwidth]{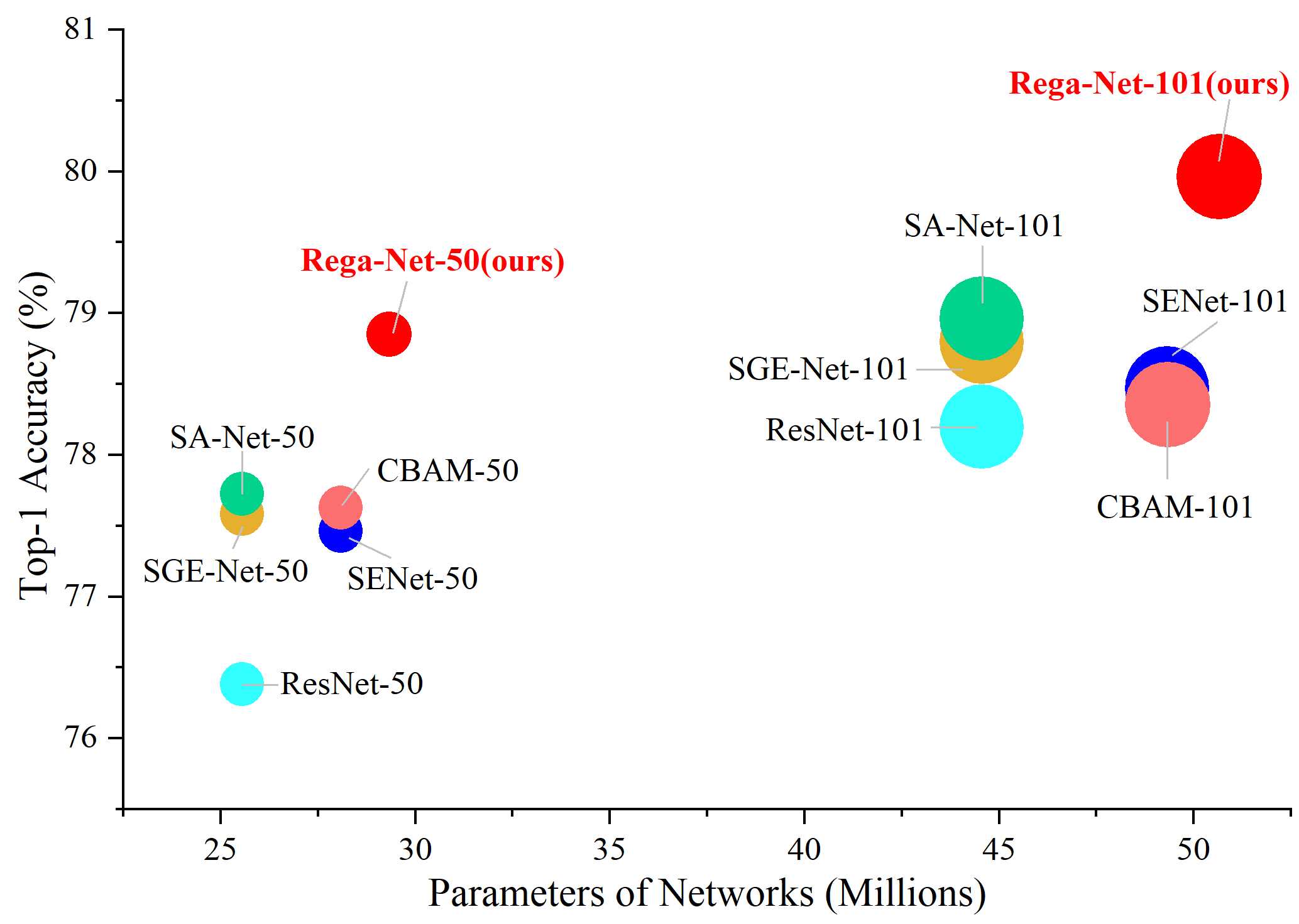}}
	\caption{Comparisons of recently SOTA attention models on ImageNet-1k \cite{ref19}, including SENet \cite{ref20}, CBAM \cite{ref21}, SGE-Net \cite{ref22}, SA-Net \cite{ref23}, and Rega-Net, using ResNets as backbones, in terms of accuracy, network parameters, and GFLOPs. “-50” means ResNet-50 and “-101” means ResNet-101. The size of the circles indicates the GFLOPs. Clearly, the proposed Rega-Net achieves higher accuracy while having less model complexity.}
	\setlength{\belowcaptionskip}{2cm} 
	\vspace{-1.8em} 
\end{figure}

In convolutional neural networks, we use large numbers of filters with various properties to extract features. These filters perform a sliding window operation on the image. And then, we train the networks with large-scale data to obtain the model parameters that fit the dataset best. However, such an operation also has a mass of redundant features. We notice that this sliding scan of the convolutional kernel is like the scanning behavior of human eyes. Therefore, we analyze the properties of the human eye. The attention of our human eye has the advantage of high central and low peripheral resolution \cite{ref24,ref25,ref26}. Compared with the variable resolution distributed structure, the uniformly distributed values in the conventional convolution kernel contribute to feature extraction at the same level. These uniform distributed values are difficult to fully extract the features we need to focus on. Based on these above observations, we propose a novel retina-like convolution method and attention mechanism in this letter, Rega-Net. We design convolutional kernels so that weights contribute highly in the center and lowly in the periphery. In general, during network training, we cannot artificially interfere with the distribution of functions in convolutional kernels. Filling the parameters obtained by training with standard initialization methods in this retina-like structure is unreasonable. So we came up with the Gabor function \cite{ref27,ref28,ref29}, which is similar to the human vision mechanism. We make convolutional kernel masks in imitation of the human eye variable resolution property and sample it over the distribution of the Gabor function. Finally, we fill the sampled values as the weights of convolutional kernels. In conventional convolution, if the values of kernel edges are missing, the network becomes less capable of detecting objects at the edges. Thus, we make this convolution into a feature attention module. This approach adds additional Retina-like Gabor features extracted by central kernels without changing the original feature extraction of the convolutional neural network. The contributions of this article are mainly in three aspects:

1) We propose a novel structure of circular convolutional kernels combined with the properties of the retina-like variable resolution. These retina-like convolutional kernels allow essential information to be more visible in the center position of the receptive field. At the same time, we make the retinal convolutional kernel parameters follow the Gabor function distribution. This architecture expands the receptive field of the convolutional neural network (CNN) when extracting features.

2) We design the above retina-like convolutional kernel as a retina attention structure, which is capable of extracting deeper features as well as multi-scale features. Through experimental validation on ImageNet-1k \cite{ref19}, and MS COCO \cite{ref30} datasets, we demonstrate that this structure achieves higher accuracy compared to the conventional attention module.

3) Our proposed Retina Gabor attention method is a plug-and-play module that can be applied to various deep learning tasks, such as image classification, object detection and recognition, semantic segmentation, etc.
\vspace{-0.4cm}	

\section{Related works}

\textbf{Gabor Filtert}. The Gabor filter is similar to the human visual system in frequency and directional characteristics. There is also a large body of research work on Gabor function in the field of computer vision \cite{ref27,ref31,ref32}. The calculation process of the Gabor function operation of the image is shown in Eq.(1) .
\begin{equation}
	g(x,y,\omega ,\varphi ,\sigma ) = \exp ( - \frac{{{{x'}^2} + {{y'}^2}}}{{2{\sigma ^2}}})\exp (i(\omega x'{\rm{ + }}\varphi ))
\end{equation}
\begin{equation}
	x^\prime = xcos\theta + ysin\theta, y^\prime = -xcos\theta + ycos\theta
\end{equation}

Where $(x,y)$ is the spatial position of the pixel on the image. Here, we only use the real part of the function as shown in Eq. (4). $\omega$ is the central angular frequency of a sinusoidal plane wave, $\theta$ is the anti-clockwise rotation of the Gabor function (the orientation of the Gabor filter), $\sigma$ is the sharpness of the Gabor function along with both $x$ and $y$ directions. Regarding the calculation, we treat it similarly as \cite{ref28}. Normally, we take $\theta=\pi/\omega$. And $\phi$ follow the distribution $U(0,\pi)$.
\begin{equation}
	g(x,y,\omega,\varphi,\sigma)=\exp (-\frac{{{{x'}^2} + {{y'}^2}}}{{2{\sigma^2}}})\exp(\omega x'{\rm{+}}\varphi)\label{pythagorean}
\end{equation}
\begin{equation}
	\omega_n=\frac{\pi}{2}{\sqrt2^{-(n-1)}},\theta_m =\frac{\pi}{8}(m-1)
\end{equation}

Where, $n=1,2,...,5$,$m=1,2,...8$.
\vspace{-0.4cm}

\section{Proposed Method}
\label{sec:guidelines}

\subsection{Rega Kernel}
Inspired by the non-uniform sampling of the human eye, we adopt a retina-like design for the structure of convolutional kernels and propose the design idea of the Rega kernel. As shown in Fig.2, we first generate a non-uniform numerical mask $\bf{M}$. This structure is capable of forming a circular-like mask. The size of the mask is 7$\times$7, ${\bf{M}}\in \mathbb{R}^{7\times7} $. Depending on the circle's radius, we fill convolutional kernels with 0 or 1. When the mask value is 1, the original convolutional kernel value is retained in that position. When the mask value is 0, the convolutional kernel's effect at that position is reduced or removed. In Fig. 3, we refer to the value of the convolutional kernel that satisfies the condition ${r_1}<r\le{r_2}$ as the One-gate Activation Point (OAP), which denotes the sampling point closest to the edge. The sampled points of the convolutional kernel, like the OAP, contribute very little value to the overall feature maps. The sampling points that satisfy the condition $r\le{r_1}$ are called Two-gate Activation Points (TAP). After one round of filtering, these activation points are sampled for more important information. Where $r$ is the distance from the coordinates of the center point of the surrounding points, $r_2$ is 1/2 the size of the convolutional kernel, and $r_1$ is the distance of the inner layer. The middle position of the convolutional kernel, which we call Fovea Point (FP), indicates the point that contributes most to the feature sampling of the feature map. We call FP, TAP, and OAP activation points. This structure is similar to the human retina variable-resolution structure. This design differs from the dilated convolution \cite{ref33,ref34}. The contributions of dilated convolutional kernels are homogeneous. And this uniform sampling increases the size of the receptive field. Our proposed Rega kernel retains the advantage of the increased receptive field, like dilated convolution, while aggregating the information in convolutional kernels. The values in retina-like masks are calculated as shown in Eq. (5).
\begin{figure}
	\centerline{\includegraphics[width=\columnwidth]{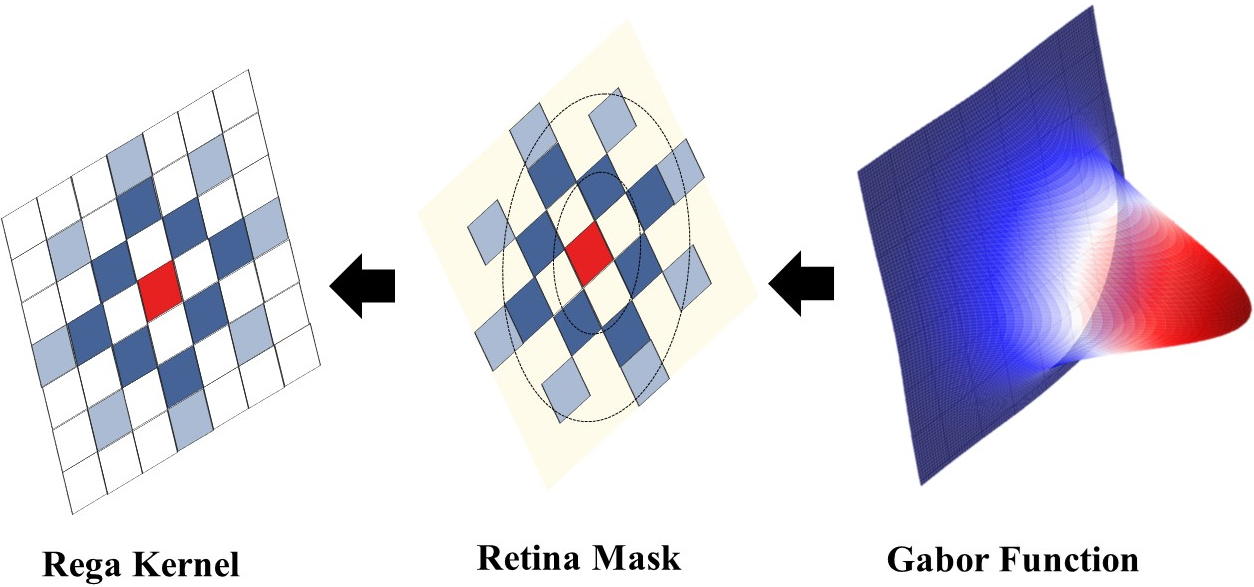}}
	\caption{The structure of the Rega kernel.}
	\setlength{\belowcaptionskip}{2cm} 
	\vspace{-2.0em} 
\end{figure}
\begin{equation}
	{{\bf{M}}_{i,j}} = \left\{ \begin{gathered}
		1,\quad \hfill {{r_1} < r \le {r_2}}\quad\\
		0,\quad  \hfill  otherwise.\\ 
	\end{gathered}  \right.
\end{equation}

Where ${{\bf{M}}_{i,j}}$ is the values of the retina mask at the position $(i,j)$. We have two considerations in the design of the retina-like mask. As shown in Fig.3(a), we set all points other than activation points to $0$. This motivation is to remove the influence of non-activation points. Thus, the interference of weakly correlated features is removed during training. In Fig.3(b), we set the values of the non-activation points to $1$. The values of the original positions are preserved in this way. So this method is a soft enhancement, without hard pruning like Fig.3(a). However, this structure in Fig.3(b) also brings some disadvantages. For instance, the number of FLOPs to calculate the parameters during the training process will increase, and the gradient calculation will be more complicated. Taking into account the above factors, we design the convolutional kernel masks in this work's manner of Fig.3(a).
\begin{figure}
	\centerline{\includegraphics[width=\columnwidth]{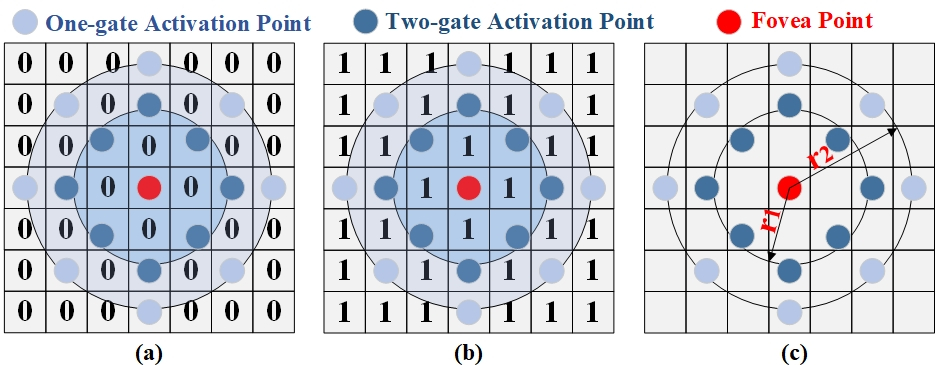}}
	\caption{The structure of the Retina masks.}
	\setlength{\belowcaptionskip}{-2cm} 
	\vspace{-1.8em} 
\end{figure}

In the above, we have completed the design of Retina masks. Then, we consider filling the mask with trainable parameters. The structural parameters of retina masks (like $0,1$) are not involved in the gradient calculation in this work. We have two ways to fill values in Retina masks. One way is to initialize the trainable parameters with random initialization. The other way is to let parameters follow a specific distribution. As shown in Fig.2, we follow the sampling rule of the Retia mask to pick up the points for filling in the convolutional kernels, which combines Gabor functions with Retina masks. It is worth noting that the parameters of the Gabor function we used are all trainable parameters. We refer to \cite{ref27} for the generation of Gabor convolutional kernels. Suppose the feature map of the input Gabor convolutional kernel is $F_{in}$, and the size of the input feature map channel is $C_{in}$. The output feature map after convolution calculation is $F_{out}$, and the size of the output feature map channel is $C_{out}$. The generated Gabor convolutional kernel is shown in Eq.(6).
\begin{equation}
	{\bf{K}} = g(x,y,\omega ,\varphi ,\sigma )
\end{equation}

Where ${\bf{K}}\in\mathbb{R}^{C_{in}\times C_{out}\times7\times7}$. The parameters of the Gabor kernel are all learnable parameters that can be used in the gradient calculation when we train the model. After the Gabor kernel dot with the Retina mask, we obtain the final Retina Gabor convolutional kernel, as shown in Eq.(7).
\begin{equation}
	{\bf{\hat K}} = {\bf{K}} \otimes {\bf{M'}}
\end{equation}

Where $\bf{M'}$ is initialized by copying from $\bf{M}$ according to channel, ${\bf{M'}}\in\mathbb{R}^{C_{in}\times C_{out}\times7\times7}$, ${\bf{\hat K}}$ is the values of kernel after retina-like sampling. $\otimes$ denotes element-wise multiplication.

\vspace{-1.1em}

\subsection{Rega Attention Network}

The structure of our designed Rega network is shown in Fig.4(a). Here, we take ResNet as the base model to illustrate the structure. We enhance the output feature maps $\bf{F}_{C1}$ and $\bf{F}_{C2}$ in $C1$  and $C2$ layers of ResNet by Rega attention module, respectively. The structure of Rega attention is shown in Fig. 4(b). We call the combination of “RG Conv” and “BN+ReLU” Retina Gabor (RG) Blocks. Moreover, the number of RG blocks is selective in the Rega attention module. The stronger the feature, the greater the number. The calculation of Rega attention is shown in Eq. (8).
\begin{equation}
	\begin{aligned}
		{{\bf{R}}_a}({{\bf{F}}_{{C_i}}}) &= \sigma (AvgPool(RegaConv({{\bf{F}}_{{C_i}}},{\bf{\hat K}})))\\
		&= \sigma (AvgPool({{{\bf{\hat F}}}_{{C_i}}}))
	\end{aligned}
\end{equation}
\begin{equation}
	{{\bf{\hat F}}_{{C_i}}} = RegaConv({{\bf{F}}_{{C_i}}},{{\bf{\hat K}}_i}) = \sum\limits_{n = 1}^N {{\bf{F}}_{{C_i}}^{(n)} \otimes {\bf{\hat K}}_i^{(n)}}
\end{equation}

Where ${{\bf{\hat F}}_{{C_i}}}$ is the feature map obtained after the $n{\rm{-th}}$ convolutional kernel operation, ${C_i}$ is the number of channels of the input feature map and ${{\bf{F}}_{{C_i}}}$ is the input feature map of the ${C_i}$ layer, ${{{\bf{F}}_{{C_i}}}}\in\mathbb{R}^{C_i\times H_i\times W_i}$.   $H_i$ is the height of the feature map, and $W_i$ is the width of the feature map. $\sigma$ denotes the sigmoid function. $AvgPool$ is the average pooling layer. We use $AvgPool$ to change the size of output feature maps. ${{\bf{\hat F}}_{{C_i}}}$ is the $C_i$ layer for Retina convolution operation. ${{\bf{R}}_a}({{\bf{F}}_{{C_i}}})$ denotes the attentional feature matrix obtained after the Rega attention operation. The final output of the attention feature maps is calculated as shown in Eq. (10).

\begin{equation}
	\begin{aligned}
		{{\bf{R}}_{out}}({{\bf{F}}_{{C_i}}}) & = {{\bf{F}}_{{C_i}}} \otimes {{\bf{R}}_a}({{\bf{F}}_{{C_i}}})\\
		& = {{\bf{F}}_{{C_i}}} \otimes \sigma (AvgPool({{{\bf{\hat F}}}_{{C_i}}}))
	\end{aligned}
\end{equation}

Where ${{\bf{R}}_{out}}({{\bf{F}}_{{C_i}}})$ is the attention feature maps matrix of the layer $C_i$. In the structure of Fig.4(a), we adopt a skipped layer of residual connections. We input the feature maps of $C1$ layer and $C2$ layer into the Rega attention module and obtain the attention feature maps of layers $C1$ and $C2$ respectively. Then we concatenate ${{\bf{R}}_{C1}}({{\bf{F}}_{C1}})$ and ${{\bf{R}}_{C2}}({{\bf{F}}_{C2}})$ with ResNet block's final $C4$ layer output feature maps. Finally, the 1$\times$1 convolution operation ($Conv_{1\times1}$) is used to integrate the final output channels to the same size as $C4$. The operation is shown in Eq. (11), where ${{\bf{F}}_{output}}$ is the final output feature map, ${{{\bf{F}}_{{output}}}}\in\mathbb{R}^{C_4\times H_4\times W_4}$.

\begin{small}
	\vspace{-1.8em} 
	\begin{equation}
		{{\bf{F}}_{output}} = Con{v_{1 \times 1}}(concat[{{\bf{R}}_{C1}}({{\bf{F}}_{C1}}),{{\bf{R}}_{C2}}({{\bf{F}}_{C2}}),C4])
	\end{equation}
	\vspace{-0.7cm}
	
\end{small}
\begin{figure}
	\centerline{\includegraphics[width=\columnwidth]{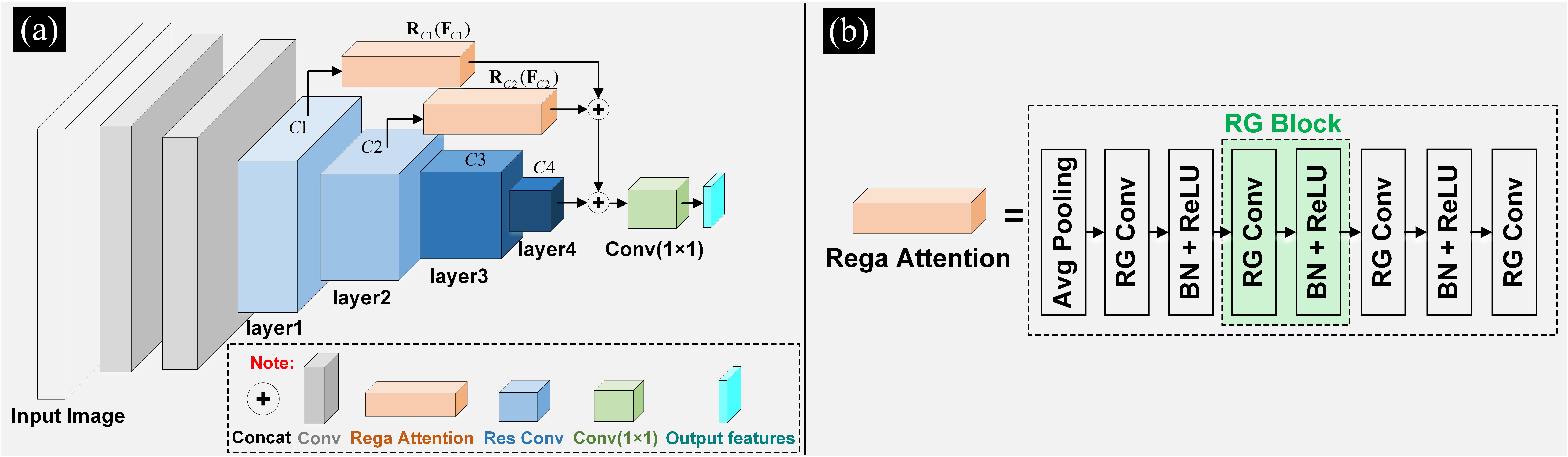}}
	\caption{The structure of the Rega-Net. “Conv” means convolution operations, and “Avg Pooling” means average pooling. “RG Conv” means Retina Gabor Convolution operation, “BN” means Batch Normalization, and “RG Block” means Retina Gabor Block that includes a “RG Conv” operation and a “BN+ReLU” operation.}
	\setlength{\belowcaptionskip}{2cm} 
	\vspace{-1.8em}
\end{figure}

\section{Experiments}

\subsection{Implementation Details}
In the experiments, we evaluate Rega-Net on the standard benchmarks: ImageNet-1k for classification and MS COCO 2017 for object detection and recognition. To ensure the fairness of the experiments, we chose the PyTorch framework to evaluate all experiments.

\textbf{Dataset}. Our image classification experiments are all performed on the ImageNet-1K dataset, which contains 1.28 M training images and 50k validation images from 1000 classes. We conduct all object detection and recognition experiments on the challenging MS COCO 2017 dataset that includes 80 object classes. Following the common practice, we use all 115K images in the trainval35k split for training and all 5K images in the minival split as validation for the analysis study.

\textbf{Experiment Setup}. We implement our networks with Python 3.8 and PyTorch 1.8.0. The Rega-Net and benchmark models’ training is conducted on 4 Geforce RTX 3080Ti GPUs. For the classification task, we set the learning rate initially as 0.01 and decreased it by a factor of 10 after every 30 epochs for 100 epochs in total. The optimization is performed using the stochastic gradient descent (SGD) with a weight decay of 1e-4, the momentum is 0.9, and the batch size is 16 per GPU. We train networks on the training set and report the Top-1 and Top-5 accuracies on the validation set with a single 224$\times$224 central crop. The learning rate for object detection and recognition tasks is 1e-4. And we choose the MultiStepLR scheduler for the learning rate. The AdamW is used with a weight decay of 1e-3, the momentum is 0.9, and the batch size is 2 per GPU within 12 epochs. We follow the standard set of evaluating object detection via the standard mean Average Precision (AP) scores at different box IoUs or object scales, respectively.
\vspace{-1.2em}

\subsection{Ablation Study}
For ablation study tasks, the structure we used is shown in Fig. 5. To reduce the network's complexity, we do not use a multi-scale feature fusion structure. Therefore, our model reaches convergence in a relatively short time.
\begin{figure}
	\centerline{\includegraphics[width=\columnwidth]{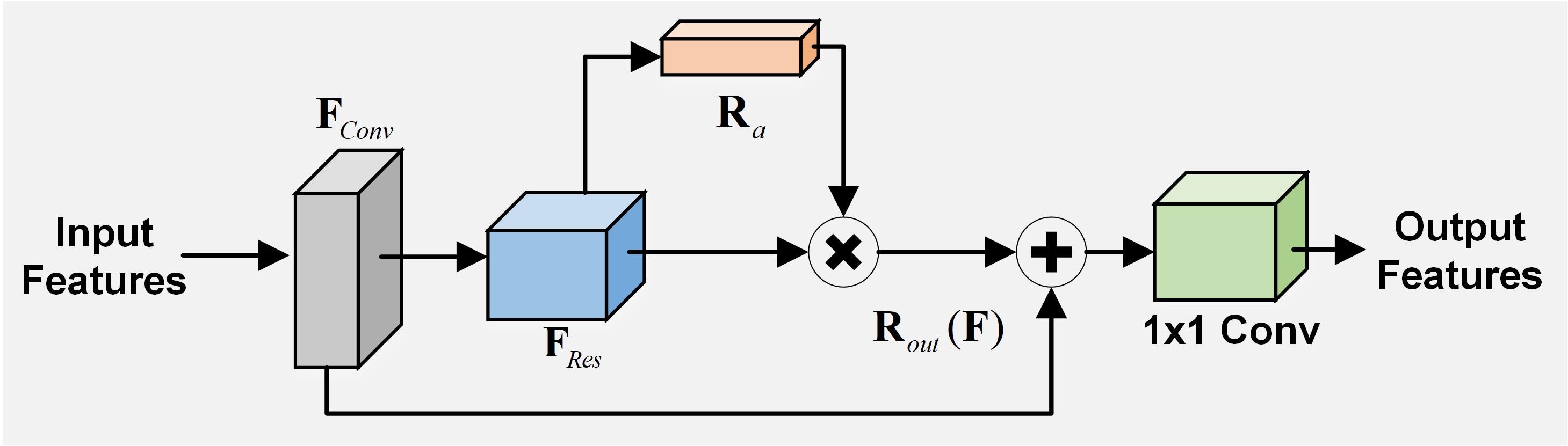}}
	\caption{The structure of Single-structure Rega attention module. We do not use a multi-scale feature fusion structure for ablation study.}
	\setlength{\belowcaptionskip}{2cm} 
	\vspace{-0.4cm}
\end{figure}

To verify the effectiveness of our designed Rege attention module, we first trained it in four residual blocks of ResNet-50, Layer1, Layer2, Layer3, and Layer4. And we placed the Rega attention block after each block and tested it on the ImageNet-1K val dataset, and the results are shown in Table I.

From Table 1 we summarise that when we add the Rega attention block to Layer4, the test accuracy is highest on the ImageNet-1K validation dataset. And the accuracy can be increased by at most 2.468\% compared to the original without Rega attention block. Therefore, in the following experiments, we prefer to add the Rega attention block to the last layer of feature map extraction for feature enhancement. 

\begin{table}
	\caption{Results of ablation experiments performed on the ResNet-50 network.}
	\vspace{-0.2cm}
	\label{table}
	\scriptsize
	\setlength{\tabcolsep}{3pt}	
	\centering
	\begin{tabular}{cccc|c|c} 
		\hline
		\toprule[0.2mm]
		Layer1       & Layer2      & Layer3       & Layer4       & Top-1 Acc (\%)     & Top-5 Acc (\%)  \\ 
		\hline
		\toprule[0.2mm]
		\checkmark   &             &              &              & 78.012             & 93.542          \\
		& \checkmark  &              &              & 78.231             & 93.721          \\
		&             & \checkmark   &              & 78.623             & 93.821          \\
		&             &              & \checkmark   & \textbf{78.852}    & \textbf{94.12}  \\ 
		\hline
		\toprule[0.2mm]
	\end{tabular}
	\vspace{-2.5em} 
\end{table}
\vspace{-1.2em}

\subsection{Classification on ImageNet-1k}
We conduct classification experiments on the ImageNet-1k dataset. The baseline we choose is ResNet-50 and ResNet-101. We compare our Rega-Net with some SOTA attention modules. And we choose the evaluation metrics with GFLOPs, Parameters, and accuracy (Top-1 and Top-5 accuracy). As shown in Table II, Rega-Net almost has the same parameters but achieves 1.128\% gains in Top-1 accuracy and 0.332\% improvement in Top-5 accuracy (on ResNet-50) with SA-Net. When using the ResNet-101 backbone, compared with SOTA attention modules, Rega-Net has 1.02\% accuracy (Top-1) improvement with SENet \cite{ref20}. Compared with SA-Net \cite{ref23}, Rega-Net has 1.003\% gains on Top-1 accuracy and 1.06\% gains in terms of Top-5 when we choose ResNet-101 as the backbone.

\begin{table}
	\caption{Comparisons of different attention methods on ImageNet-1k.}
	\vspace{-0.2cm}
	\label{table}
	\scriptsize
	\setlength{\tabcolsep}{3pt}		
	\centering
	\begin{tabular}{c|c|c|c|c|c} 
		\hline
		\toprule[0.2mm]
		Attention Methods       & Backbones                          & Param.  & GFLOPs & Top-1 Acc (\%) & Top-5 Acc (\%)  \\ 
		\hline
		\toprule[0.2mm]
		ResNet \cite{ref35}     & \multirow{6}{*}{ResNet-50}         & 25.557M & 4.122  & 76.384         & 92.908          \\
		SENet \cite{ref20}      &                                    & 28.088M & 4.130  & 77.462         & 93.696          \\
		CBAM \cite{ref21}       &                                    & 28.090M & 4.139  & 77.626         & 93.662           \\
		SGE-Net \cite{ref22}    &                                    & 25.559M & 4.127  & 77.584         & 93.664          \\
		SA-Net \cite{ref23}     &                                    & 25.557M & 4.125  & 77.724         & 93.798          \\
		Rega-Net(\textbf{Ours}) &                                    & 29.325M & 4.230  & \textbf{78.852}& \textbf{94.120}  \\ 
		\hline
		ResNet \cite{ref35}     & \multirow{6}{*}{ResNet-101}        & 44.549M & 7.849  & 78.200         & 93.906          \\
		SENet \cite{ref20}      &                                    & 49.327M & 7.863  & 78.468         & 94.102          \\
		CBAM \cite{ref21}       &                                    & 49.330M & 7.879  & 78.354         & 94.064          \\
		SGE-Net \cite{ref22}    &                                    & 44.553M & 7.858  & 78.798         & 94.368          \\
		SA-Net \cite{ref23}     &                                    & 44.551M & 7.854  & 78.960         & 94.492          \\
		Rega-Net(\textbf{Ours}) &                                    & 50.661M & 7.925  & \textbf{79.963}&\textbf {95.552} \\
		\hline
		\toprule[0.2mm]
	\end{tabular}
	\vspace{-2em} 
\end{table}
\vspace{-1.4em}

\subsection{Object Detection and Recognition}

We conduct object detection and recognition experiments on COCO 2017 benchmark. For the experiment, we reproduce FCOS \cite{ref36}, Faster R-CNN \cite{ref37}, YOLOv4 \cite{ref38}, and RetinaNet \cite{ref39} in our PyTorch framework to estimate the performance improvement of Rega-Net. The experimental results are summarized in Table III. We can see that Rega-Net improves the accuracy compared with SENet and SA-Net. We use mean AP (mAP) over different IoU thresholds from 0.5 to 0.95 for evaluation. We choose ResNet-50 and CSPDarknet-53 as the backbone. In the design of the comparison experiments, we first obtained the accuracy of the baseline model (without attention) by training. Next, we added SENet, SA-Net, and RegaNet to the backbone and trained to obtain the accuracy of each of the four detectors.

\begin{table}
	\caption{Object detection results of different attention methods on COCO val2017.}
	\vspace{-0.2cm}
	\label{table}
	\scriptsize
	\setlength{\tabcolsep}{3pt}	
	\centering
	\begin{tabular}{c|c|c|c|c|c|c|c} 
		\hline
		\toprule[0.2mm]
		Backbones        & Detectors                               & AP50:95       & AP50          & AP75          & APS           & APM           & APL  \\ 
		\hline
		\toprule[0.2mm]
		ResNet-50        & \multirow{4}{*}{FCOS \cite{ref36}}      & 34.6          & 51.3          & 36.1          & 14.3          & 36.4          & 41.5 \\
		+ SENet          &                                         & 35.4          & 52.2          & 36.6          & 15.1          & 36.8          & 41.5 \\
		+ SA-Net         &                                         & 36.5          & 53.4          & 37.5          & 15.3          & 37.6          & 42.3 \\
		+ Rega-Net(\textbf{Ours})  &                               & \textbf{37.8} & \textbf{54.6} & \textbf{37.6} & \textbf{15.4} & \textbf{37.8} & \textbf{43.5} \\ 
		\hline
		ResNet-50 & \multirow{4}{*}{Faster R-CNN \cite{ref37}}     & 36.4          & 58.4          & 39.1          & 21.5          & 40.0          & 46.6 \\
		+ SENet          &                                         & 37.7          & 60.1          & 40.9          & 22.9          & 41.9          & 48.2 \\
		+ SA-Net         &                                         & 38.7          & 61.2          & 41.4          & 22.3          & 42.5          & 49.8 \\
		+ Rega-Net(\textbf{Ours})  &                               & \textbf{39.9} & \textbf{62.3} & \textbf{42.3} & \textbf{24.6} & \textbf{43.5} & \textbf{49.9} \\ 
		\hline
		CSPDarknet-53    & \multirow{4}{*}{YOLOv4 \cite{ref38}}    & 41.2          & 62.8          & 44.3          & 24.3          & 46.1          & 55.2 \\
		+ SENet          &                                         & 42.0          & 63.4          & 45.2          & 24.9          & 46.8          & 55.7 \\
		+ SA-Net         &                                         & 42.6          & 64.2          & 45.8          & 23.1          & 45.5          & 55.6 \\
		+ RegaNet(\textbf{Ours})  &                                & \textbf{43.1} & \textbf{64.6} & \textbf{45.9} & 24.8          & 46.8          & \textbf{55.9} \\ 
		\hline
		ResNet-50        & \multirow{4}{*}{RetinaNet \cite{ref39}} & 35.6          & 55.5          & 38.3          & 20.0          & 39.6          & 46.8 \\
		+ SENet          &                                         & 36.0          & 56.7          & 38.3          & 20.5          & 39.7          & 47.7 \\
		+ SA-Net         &                                         & 37.5          & 58.5          & 39.7          & 21.3          & 41.2          & 45.9 \\
		+ Rega-Net(\textbf{Ours}) &                                & \textbf{38.6} & \textbf{58.9} & \textbf{40.9} & 20.4          &\textbf{42.1}  &\textbf{48.6} \\
		\hline
		\toprule[0.2mm]
	\end{tabular}
	\vspace{-2.5em} 
\end{table}

\section{Conclusion}
We propose a novel method for designing convolutional kernels based on the retina-like principle in this letter. And we design a state-of-the-art attention module named Rega-Net. Experimental results show that the proposed method increases Top-1 accuracy by up to 2.468\% on image classification compared to the original network. The mAP is increased by up to 3.5\% on object detection. The accuracy of CNN is effectively improved when compared with SOTA networks. However, Rega-Net still needs further optimization in terms of speed and computational complexity.



\begin{thebibliography}{34}
	\setcounter{enumiv}{21}
	\begin{spacing}{0.81}    	
		\bibitem{ref1}X. Zeng, W. Wu, G. Tian, F. Li, and Y. Liu, ``Deep superpixel convolutional network for image recognition,'' {\em IEEE Signal Process. Lett.}, vol.28, pp.922-926, 2021.    
		\bibitem{ref2}Z. Liu, Y. Lin, Y. Cao, H. Hu, Y. Wei, Z. Zhang, S. Lin, and B. Guo, ``Swin transformer: Hierarchical vision transformer using shifted windows,'' in {\em Proc. IEEE Int. Conf. Comput. Vision}, Virtual, Online, Canada, 2021, pp.10012-10022.    
		\bibitem{ref3}Z. Liu, H. Mao, C. Wu, C. Feichtenhofer, T. Darrell, and S. Xie, ``A convnet for the 2020s,'' 2022, {\em arXiv:2201.03545}.	
		\bibitem{ref4}F. Akyon, S. Altinuc, and A. Temizel, ``Slicing aided hyper inference and finetuning for small object detection,'' 2022, {\em arXiv:2202.06934}.		
		\bibitem{ref5}M. Guo, C. Lu, Z. Liu, M. Cheng, and S. Hu, ``Visual attention network,'' 2022, {\em arXiv:2202.09741}.			
		\bibitem{ref6}S. Zhao, L. Zhang, Y. Shen, and Y. Zhou, ``Refinednet: a weakly supervised refinement framework for single image dehazing,'' {\em IEEE Trans. Img. Proc.}, vol.30, pp.3391-3404, Mar.9, 2021.			
		\bibitem{ref7}X. Liu, Y. Ma, Z. Shi, and J. Chen, ``Griddehazenet: Attention-based multi-scale network for image dehazing,'' in {\em Proc. IEEE Int. Conf. Comput. Vision}, Los Alamitos, CA, USA, 2019, pp.7314-7323.	
		\bibitem{ref8}S. Kim and Y. Hwang, ``A survey on deep learning based methods and datasets for monocular 3d object detection,''{\em Electronics}, vol.10, no.4, pp.517, Feb. 2021.		
		\bibitem{ref9}H. Shuai, X. Xu, and Q. Liu, ``Backward attentive fusing network with local aggregation classifier for 3d point cloud semantic segmentation,'' {\em IEEE Trans. Img. Proc.}, vol.30, pp.4973-4984, May.14, 2021.
		\bibitem{ref10}J. Guo, X. Xing, W. Quan, D. Yan, Q. Gu, Y. Liu, and X. Zhang, ``Efficient center voting for object detection and 6d pose estimation in 3d point cloud,'' {\em IEEE Trans. Img. Proc.}, vol.30, pp.5072-5084, May.19, 2021.	
		\bibitem{ref11}Y. Guo, H. Wang, Q. Hu, H. Liu, L. Liu, and M. Bennamoun, ``Deep learning for 3d point clouds: A survey,'' {\em IEEE Trans. Pattern Anal. Mach. Intell.}, vol.43, no.12, pp.4338-4364, Nov.3, 2020.		
		\bibitem{ref12}W. Wang, Y. Cao, J. Zhang, F. He, Z. Zha, Y. Wen, and D. Tao, ``Exploring sequence feature alignment for domain adaptive detection transformers,'' in {\em Proc. ACM Int. Conf. Multimed.}, Virtual, Online, China, 2021, pp.1730-1738.	
		\bibitem{ref13}K. Oksuz, B. Cam, S. Kalkan, and E. Akbas, ``Imbalance problems in object detection: A review,'' {\em IEEE Trans. Pattern Anal. Mach. Intell.}, vol.43, no.10, pp.3388-3415, Sep.2, 2020.	
		\bibitem{ref14}X. Liang, L. Wu, J. Li, Y. Wang, Q. Meng, T. Qin, W. Chen, M. Zhang, T. Liu, ``R-drop: regularized dropout for neural networks,'' {\em Adv. neural inf. proces. syst.}, Montreal, QC, Canada, 2021.	
		\bibitem{ref15}D. Zhang, K. Ahuja, Y. Xu, Y. Wang, and A. Courville, ``Can subnetwork structure be the key to out-of-distribution generalization?'' in {\em Proc. Int. Conf. Mach. Learn.}, Virtual Only, 2021, pp.12356-12367.	
		\bibitem{ref16}A. Correia and E. Colombini, ``Attention, please! a survey of neural attention models in deep learning,'' 2021, {\em arXiv:2103.16775}.	
		\bibitem{ref17}Z. Pan, B. Zhuang, H. He, J. Liu, and J. Cai, ``Less is more: Pay less attention in vision transformers,'' 2021, {\em arXiv:2105.14217}.	
		\bibitem{ref18}Y. Li, T. Yao, Y. Pan, and T. Mei, ``Contextual transformer networks for visual recognition,'' {\em IEEE Trans. Pattern Anal. Mach. Intell.}, 2022.
		\bibitem{ref19}J. Deng, W. Dong, R. Socher, L. Li, K. Li, and F. Li, ``Imagenet: A large-scale hierarchical image database,'' in {\em Proc. IEEE Conf. Comput. Vision Pattern Recognit.}, Miami, FL, USA, 2009, pp.248-255.
		\bibitem{ref20}J. Hu, L. Shen, and G. Sun, ``Squeeze-and-excitation networks,'' in {\em Proc. IEEE Conf. Comput. Vision Pattern Recognit.}, Salt Lake City, UT, USA, 2018, pp.7132-7141.
		\bibitem{ref21}S. Woo, J. Park, J. Lee, and I. Kweon, ``Cbam: Convolutional block attention module,'' in {\em Proc. Eur. Conf. Comput. Vision}, Munich, Germany, 2018, pp.3-19.
		\bibitem{ref22}X. Li, X. Hu, and J. Yang, ``Spatial group-wise enhance: Improving semantic feature learning in convolutional networks,'' 2019, {\em arXiv:1905.09646}.
		\bibitem{ref23}Q. Zhang and Y. Yang, ``Sa-net: Shuffle attention for deep convolutional neural networks,'' {\em IEEE Int. Conf. Acoust. Speech Signal Process. Proc.}, Toronto, Ontario, Canada, 2021, pp.2235-2239.
		\bibitem{ref24}H. Yang, S. Qi, W. Chao, S. Yang, and X. Wang, ``Image analysis by logpolar exponent-fourier moments,'' {\em Pattern Recogn.}, vol.101, no.107177, May. 2020.
		\bibitem{ref25}V. Traver and A. Bernardino, ``A review of log-polar imaging for visual perception in robotics,'' {\em Robot. Autonom. Syst.}, vol.58, no.4, pp.378-398, 2010.
		\bibitem{ref26}D. Li, R. Du, A. Babu, C. Brumar, and A. Varshney, ``A log-rectilinear transformation for foveated 360-degree video streaming,'' {\em IEEE Trans. Visual. Comput. Graph.}, vol.27, no.5, pp.2638-2647, May. 2021.
		\bibitem{ref27}S. Luan, C. Chen, B. Zhang, J. Han, and J. Liu, ``Gabor convolutional networks,'' {\em IEEE Trans. Image Process.}, vol.27, no.9, pp.4357-4366, 2018.	
		\bibitem{ref28}A. Alekseev and A. Bobe, ``Gabornet: Gabor filters with learnable parameters in deep convolutional neural network,'' {\em Int. Conf. Eng. Telecommun.}, Dolgoprudny, Russia, 2019, pp.1-4.
		\bibitem{ref29}S. Meshgini, A. Aghagolzadeh, and H. Seyedarabi, ``Face recognition using gabor filter bank, kernel principle component analysis and support vector machine,'' {\em Int. J. Comput.}, vol.4, no.5, 2012.
		\bibitem{ref30}T. Lin, M. Maire, S. Belongie, J. Hays, P. Perona, D. Ramanan, P. Doll´ar, and C. Zitnick, ``Microsoft coco: Common objects in context.'' in {\em Proc. Eur. Conf. Comput. Vision}, Zurich, Switzerland, 2014, pp.740-755.
		\bibitem{ref31}B. Zhang, Y. Gao, S. Zhao, and J. Liu, ``Local derivative pattern versus local binary pattern: face recognition with high-order local pattern descriptor,'' {\em IEEE Trans. Image Process.}, vol.19, no.2, pp.533-544, 2009.
		\bibitem{ref32}C. Liu and H. Wechsler. ``Independent component analysis of gabor features for face recognition,'' {\em IEEE Trans. Neural Network.}, vol.14, no.4, pp.919-928, 2003.
		\bibitem{ref33}F. Yu, V. Koltun, and T. Funkhouser, ``Dilated residual networks,'' in {\em Proc. IEEE Conf. Comput. Vision Pattern Recognit.}, Honolulu, HI, USA, 2017, pp.472-480.
		\bibitem{ref34}X. Zhen, R. Chakraborty, N. Vogt, B. Bendlin, and V. Singh, ``Dilated convolutional neural networks for sequential manifold-valued data,'' in {\em Proc. IEEE Int. Conf. Comput. Vision}, Los Alamitos, CA, USA, 2019, pp.10621-10631.
		\bibitem{ref35}K. He, X.Zhang, S. Ren, and J. Sun, ``Deep residual learning for image recognition,'' in {\em Proc. IEEE Conf. Comput. Vision Pattern Recognit.}, Las Vegas, NV, USA, 2016, pp.770-778.
		\bibitem{ref36}Z. Tian, C. Shen, H. Chen, and T. He, ``Fcos: Fully convolutional one-stage object detection,'' in {\em Proc. IEEE Int. Conf. Comput. Vision}, Seoul, Korea, 2019, pp.9627-9636.
		\bibitem{ref37}S.Ren, K. He, R. Girshick, and J. Sun, ``Faster r-cnn: Towards real-time object detection with region proposal networks,'' {\em Adv. neural inf. proces. syst.}, Cambridge, MA, USA, 2015, pp.91-99.
		\bibitem{ref38}A. Bochkovskiy, C. Wang, and H. Liao, ``Yolov4: Optimal speed and accuracy of object detection,'' 2020, {\em arXiv:2004.10934}.
		\bibitem{ref39}T. Lin, P. Goyal, R. Girshick, K. He, and P. Doll´ar, ``Focal loss for dense object detection,'' in {\em Proc. IEEE Int. Conf. Comput. Vision}, Venice, Italy, 2017, pp.2980-2988.
	\end{spacing}	
\end{thebibliography}
\end{document}